\documentclass[pmlr,twocolumn,10pt]{jmlr} 

\usepackage{longtable}

\usepackage{booktabs}
\usepackage{caption}
\usepackage{stfloats}
\usepackage{graphicx}
\usepackage{afterpage}
\usepackage{subcaption}

\usepackage{siunitx}

\usepackage[switch]{lineno}

\usepackage{comment}

\theorembodyfont{\upshape}
\theoremheaderfont{\scshape}
\theorempostheader{:}
\theoremsep{\newline}

\jmlrvolume{}
\jmlryear{2024}
\jmlrsubmitted{}
\jmlrpublished{}
\jmlrworkshop{Conference on Health, Inference, and Learning (CHIL) 2024} 

\title[Interpretable breast cancer classification]{Interpretable breast cancer classification using CNNs on \titlebreak mammographic images}

\author{%
\Name{Ann-Kristin Balve} \Email{akdb3@cam.ac.uk}\\
\addr University of Cambridge, United Kingdom
\AND
\Name{Peter Hendrix}\Email{p.h.g.hendrix@tilburguniversity.edu}\\
\addr Tilburg University, The Netherlands
}

\begin{document}

\maketitle

\begin{abstract}
Deep learning models have achieved promising results in breast cancer classification, yet their 'black-box' nature raises interpretability concerns. This research addresses the crucial need to gain insights into the decision-making process of convolutional neural networks (CNNs) for mammogram classification, specifically focusing on the underlying reasons for the CNN's predictions of breast cancer. For CNNs trained on the Mammographic Image Analysis Society (MIAS) dataset, we compared the post-hoc interpretability techniques LIME, Grad-CAM, and Kernel SHAP in terms of explanatory depth and computational efficiency. The results of this analysis indicate that Grad-CAM, in particular, provides comprehensive insights into the behavior of the CNN, revealing distinctive patterns in normal, benign, and malignant breast tissue. We discuss the implications of 
the current findings for the use of machine learning models and interpretation techniques in clinical practice.
\end{abstract}

\paragraph*{Data and Code Availability}
This paper uses the publicly available Mammographic Image Analysis Society (MIAS) dataset \citep{suckling2015mammographic} which was acquired through the Kaggle platform (\href{https://www.kaggle.com/datasets/kmader/mias-mammography}{mias-mammography dataset}). The MIAS dataset contains 322 images of full mammography scans, the corresponding labels, and region of interest (ROI) annotations. The dataset is licensed under \href{https://creativecommons.org/licenses/by/2.0/uk/}{CC BY 2.0 UK}. A preprocessed version of the MIAS dataset that was used for the current analyses along with the relevant preprocessing code is available on GitHub (\href{https://github.com/annkristinbalve/Interpretable_Breast_Cancer_Classification}{link}). 

\paragraph*{Institutional review board (IRB)}

The research presented here did not require IRB approval.

\section{Introduction}
\label{sec:intro}

\begin{figure*}[htbp!]
    \floatconts
    {fig:mammogram_drawio}
     {\captionsetup{aboveskip=-7.5pt}
    \caption{Proposed System: A CNN-based Mammography Classification Framework augmented with Grad-CAM for enhanced explainability.}}
    {\includegraphics[width=0.9\linewidth]{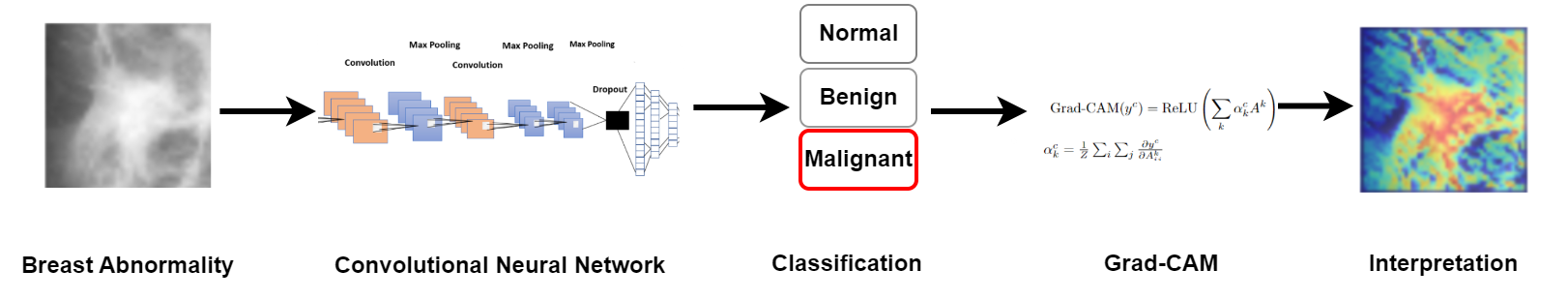}}
    \vspace*{-1.5em}
\end{figure*}

Breast cancer is the most prevalent cancer in women, and early detection through various methods, including mammography, is crucial to decrease mortality rates \citep{gco,mughal2018novel}. Mammography involves taking X-ray images of breast tissue to identify abnormalities such as calcifications and masses \citep{nci}. These breast abnormalities can be benign, non-cancerous abnormal growths not spreading outside the breast or malignant, cancerous tumors that can spread to other organs. The accurate classification of benign or malignant breast lumps is crucial due to the health risks of malignancy, yet, it is even challenging for skilled radiologists. In fact, false negatives within mammography screening can be attributed, among other factors, to human perception errors \citep{ekpo2018errors} and misinterpretations \citep{palazzetti2016analysis}. This underscores the potential of computer-aided diagnosis (CAD) and deep learning systems in assisting radiologists to more accurately interpret mammographic images, thereby potentially mitigating these diagnostic errors \citep{lee2013cognitive, coccia2020deep}.

\color{black}
Deep learning (DL) algorithms can automatically extract features and effectively represent high-dimensional data. Convolutional neural networks (CNNs), types of DL models, are particularly suitable for image recognition tasks. They have shown promising results in classifying breast cancer from screening mammograms \citep{el2021malignant, li2019benign}. Enhancements in CNN performance involve various strategies, such as pretraining on image patches extracted from the regions of interest (ROI) in mammographic images containing breast abnormalities \citep{shen2019deep}. This approach demonstrated improved model performance when subsequently training on full mammograms without ROI annotations. Further, transfer learning across different mammographic datasets, as well as using pretrained models on ImageNet, such as InceptionV3 \citep{szegedy2016rethinking}, has been shown to outperform models trained from scratch for detecting breast masses  \citep{agarwal2019automatic}. To address overfitting concerns, data augmentation - which involves creating new training samples - has been employed to enhance model robustness \citep{li2019benign}. 
    
DL, while advancing the accuracy of diagnostic models, comes at the cost of interpretability, also known as the 'black box' problem. This issue arises as deep learning uses higher-level abstractions in deep hidden layers, which, although increasing the model accuracy, make the reasoning behind its decisions more opaque to human understanding. As a response to a lack of model interpretability, a new branch in artificial intelligence (AI) research has focused on explainable AI (XAI) which aims to provide insight into the model predictions to make deep learning models more interpretable.

Recent research has explored the interpretability of neural networks, using numerical datasets such as the Wisconsin dataset which provides features describing breast abnormalities extracted from mammograms \citep{karatza2021interpretability, hakkoum2021assessing}. These studies used various methods to enhance model transparency, such as Shapley values \citep{karatza2021interpretability}, feature importance, and LIME \citep{hakkoum2021assessing}. However, with mammography relying on image analysis, the interpretability of CNNs that take images rather than numerical features as input is paramount. For instance, \citet{zhang2018interpretable} developed interpretable end-to-end CNNs that encode semantic information for each filter, while \citet{chen2019looks} proposed ProTopNet, a model that highlights image parts that motivated the model prediction by comparing prototypical image parts of a class.

However, while recent studies have focused on improving model accuracy in diagnosing breast cancer using mammographic images, the crucial aspect of model explainability has been overlooked. Although interpretable breast cancer models have been applied to numerical breast cancer datasets, the post-hoc explainability of CNNs classifying mammographic images has, to the best of our knowledge, not been explored yet. As end-to-end intrinsically interpretable DL models might trade off accuracy, this underscores the importance of post-hoc interpretability of established image classifiers, such as CNNs, to enhance transparency in DL-driven breast cancer classification.

This research seeks to bridge the gap in explainable AI (XAI) within mammography, emphasizing the critical yet often neglected aspect of explainability in AI-driven diagnostics. By evaluating three post-how interpretability algorithms (Kernel SHAP, LIME, and Grad-CAM) following the training of a CNN on the MIAS mammogram dataset, our work not only addresses the automated classification of normal, benign, and malignant breast tissue, but integrates post-hoc explainable AI techniques to uncover the CNN's predictive rationale (\figureref{fig:mammogram_drawio}). We highlight Grad-CAM's capability to reveal distinct patterns between breast abnormalities, emphasizing the potential of explainable AI in fostering trust and transparency in CNN-based mammogram classification. Our model offers deeper insights for automated classification, aiding radiologists with a transparent tool to improve clinical decision making, potentially enhancing patient outcomes and reducing diagnostic errors.



\section{Methods}

\begin{figure}[t!]
\floatconts
  {fig:subfigex2}
  {\captionsetup{aboveskip=-5pt}
  \caption{MIAS mammograms with ROI annotations.}}
  {%
    \subfigure[Normal][c]{\label{fig:normal}%
      \includegraphics[width=0.23\linewidth]{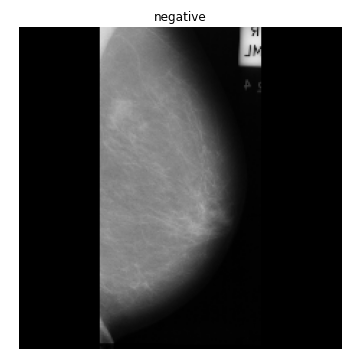}}%
    \qquad
    \subfigure[Benign][c]{  \label{fig:benign}%
      \includegraphics[width=0.23\linewidth]{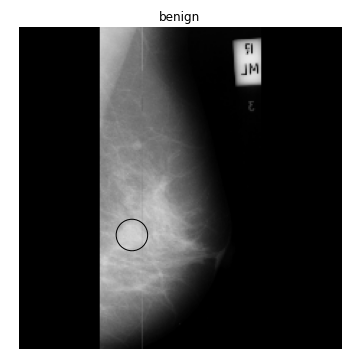}}
    \qquad
      \subfigure[Malignant][c]{\label{fig:malignant}%
      \includegraphics[width=0.23\linewidth]{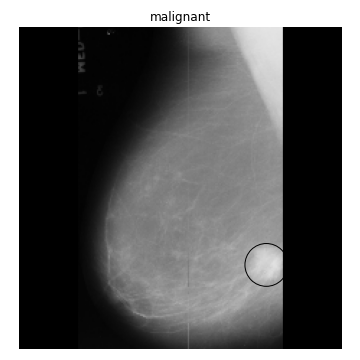}}%
  }
  \label{fig:unique_img}
  \vspace{-1em} 

\end{figure}

\begin{figure*}[htbp]
\floatconts
  {fig:preprocessing}
  {\captionsetup{aboveskip=-5pt}
  \caption{Preprocessing procedure for all images, visualized for an example image (idx:183), depicting the original image (a), a preprocessed versions after noise removal (b), image enhancement (c), extractions of the Region of Interest (ROI) (d), 90° rotation (e), vertical flipping (f), and brightness and contrast changes (g).}}
  {%
    \subfigure[Image]
    {\label{fig:orig_183}
      \includegraphics[width=0.13\linewidth]{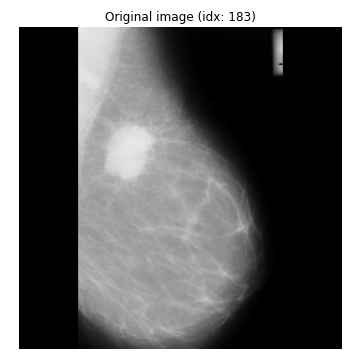}}
      \hfil 
    \subfigure[Noisefree]
    {\label{fig:noisefree}
      \includegraphics[width=0.13\linewidth]{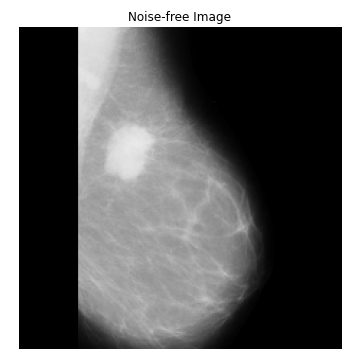}%
    }\hfil
    \subfigure[Enhanced]{%
  \label{fig:enhanced}
      \includegraphics[width=0.13\linewidth]{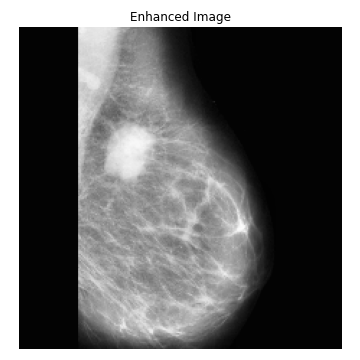}%
    }\hfil
    \subfigure[ROI]{%
      \label{fig:roi}%
      \includegraphics[width=0.13\linewidth]{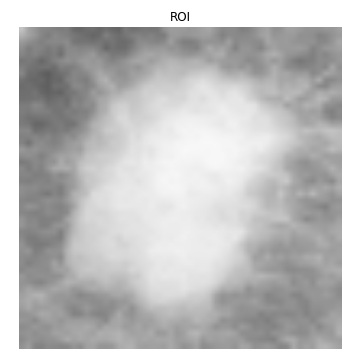}%
    }\hfil
    \subfigure[90°rotated]{%
     \label{fig:rotated}
      \includegraphics[width=0.13\linewidth]{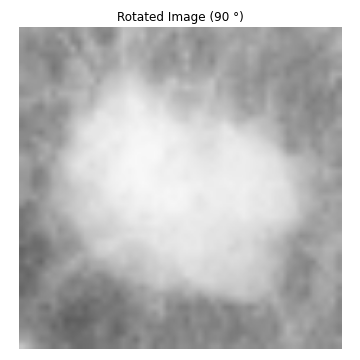}%
        }\hfil
    \subfigure[Flipped]{%
    \label{fig:flipped}
      \includegraphics[width=0.13\linewidth]{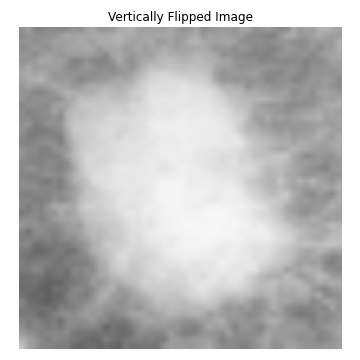}%
    }\hfil
    \subfigure[Brightness/ Contrast ]{%
  \label{fig:brightness/contrast}
      \includegraphics[width=0.13\linewidth]{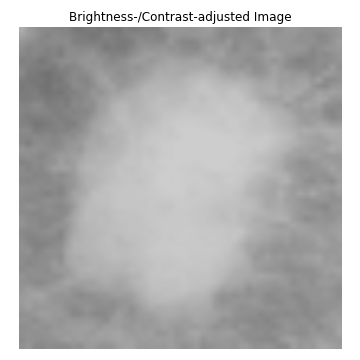}}
}
\end{figure*}

\subsection{Dataset}

The MIAS (Mammographic Image Analysis Society) dataset \citep{suckling2015mammographic} contains images of mammography scans and three corresponding labels: normal (0), benign (1), and malignant (2). Additionally, it provides the x and y coordinates and radius of the abnormality for benign and malignant images (\figureref{fig:unique_img}). The images are in PGM format and have a size of 1024 x 1024 pixels. They contain 161 pairs of RGB images resulting in 322 images in total of which 207 are normal, 64 are benign and 51 are malignant.

\subsection{Preprocessing}

The mammographic images underwent a preprocessing procedure adapted from \cite{el2021malignant}. This preprocessing of the raw images (\figureref{fig:orig_183}) included noise artifact removal (\figureref{fig:noisefree}) and image quality enhancement using contrast limited adaptive histogram equalization (CLAHE) (\cite{zuiderveld1994contrast}, (\figureref{fig:enhanced}). For abnormal images, the region of interest (ROI) was extracted using the x and y coordinates and radius of the abnormality (\figureref{fig:roi}), while normal images were cropped using a central breast area. We removed four abnormal images without ROI annotations from the dataset. After cropping, all images were resized to 224 x 224 pixels and visually inspected. Two ROI images with incorrect coordinate sets were identified and removed from the dataset, resulting in a final dataset of 316 images.

\subsection{Splitting, data augmentation, balancing} 
We split the dataset into a training (proportion of images: 0.70), validation (0.15), and test set (0.15). To ensure a representative proportion of each class in every set, we applied a stratified split. To mitigate potential overfitting we augmented the dataset by creating new samples through small transformations of the original data \citep{perez2017effectiveness,oza2022image, li2019benign}. This included rotation (0°, 90°, 180°, 270°) (\figureref{fig:rotated}), vertical flipping (\figureref{fig:flipped}) and random brightness and contrast changes in the ranges (-15,15) and (0.5,1.5) respectively (\figureref{fig:brightness/contrast}). After removing two duplicates, data augmentation resulted in a training set of 3534 images. The training dataset was subsequently balanced, yielding a dataset consisting of 1566 images for training (528 normal, 528 benign, 528 malignant images), 47 images (31 normal, 9 benign, 8 malignant) for validation, and 48 images for testing (31 normal, 9 benign, 8 malignant). 


\subsection{CNN architecture and hyperparameter tuning} 

\begin{figure*}[htbp]
\floatconts
  {fig:cnn_architecture}
  {\captionsetup{aboveskip=-15pt}
  \caption{CNN architecture adapted from \cite{el2021malignant}. ROIs are fed into a network consisting of three convolutional layers and four pooling layers, followed by a flattening layer with three fully connected layers. The output layer uses a softmax activation function to classify the ROIs as normal, benign, or malignant.}}
  {\centering
   \includegraphics[width=1\textwidth]{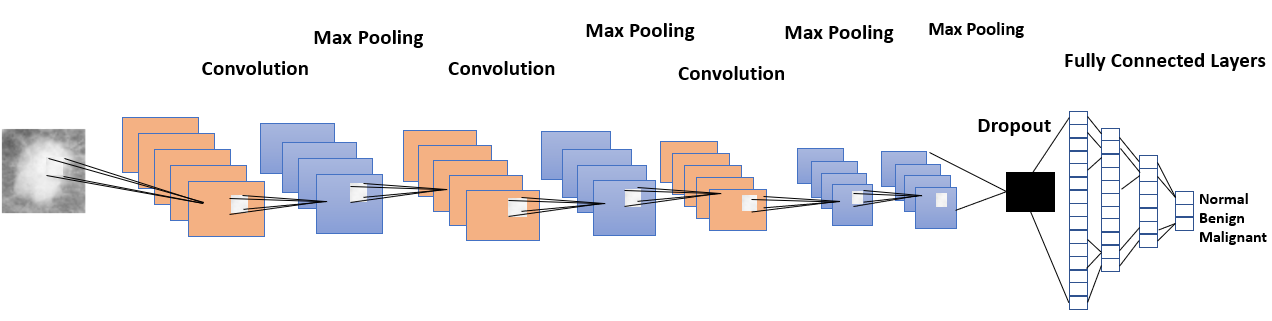}}
   \vspace{-1em} 

\end{figure*}

We fit a CNN to the data. CNNs use convolutional layers to extract features by sliding over the image and producing a feature map for features on different locations of the image. The CNN architecture used in this study was adapted from  \cite{el2021malignant}, who revealed promising results for this architecture on the MIAS dataset (accuracy: 0.95 for binary classification).  The architecture of the CNN is visualized in Figure \ref{fig:cnn_architecture} and a detailed description of it is provided in Table \ref{tab:cnn_architecture}. Prior to training, we normalised image pixels to [0,1] and one-hot encoded class labels. The learning rate and batch size were set at 0.0001 and 16, following \citet{kandel2020effect}. We used the Adam optimizer to optimize computational efficiency and minimize parameter tuning \citep{kingma2014adam}. To improve the generalization performance of the model, a dropout layer with a rate of 0.5 was applied before the last fully connected layer \citep{li2018benign}. Finally, we used the categorical cross-entropy loss function.

To improve model performance, hyperparameter tuning for the number of epochs and class weights was performed. We employed early stopping to balance the risk of overfitting, haltering training when the validation loss ceased to decrease for 10 successive epochs. The model initially trained for 50 epochs. The validation loss of the first CNN model (here referred to as 'CNN\_1') stopped improving after 14 epochs (validation loss: 0.50; validation accuracy: 0.81), thus these weights were restored and used for the model predictions. Class weight tuning was critical due to the model's sensitivity to these parameters. Besides a default uniform class weight, other weight configurations prioritizing malignant cases (e.g., 0:1, 1:2, 2:3; 0:2, 1:3, 2:4) were tested and compared. Class weights of {0:1, 1:2, 2:3} were selected, assigning higher weights to benign and malignant cases. 

\section{Analysis}

\subsection{Model Evaluation Metrics}

We evaluated the performance of the CNN using class-specific recall, precision, and F1-scores. In light of the class imbalance in the validation and test sets, averaged macro metrics and class-specific metrics (accuracy, recall, precision, F1-score, balanced accuracy) were computed. To leverage the sensitivity of model evaluation metrics to the initialization of the CNN, average performance metrics were computed over ten iterations of training and evaluation and compared against a majority class baseline \citep{Jaamour_2020}. Furthermore, the area under the receiver operating characteristics (AU-ROC) curve, which depicts the trade-off between true positive and false positive rates, was computed.

\subsection{Model Interpretability} 

To investigate the interpretability of the CNN model, three model interpretability techniques were applied: Kernel SHAP, Grad-CAM, and LIME. Below, we describe each of these techniques in more detail.

\subsubsection{SHapley Additive exPlanations (SHAP)}


SHAP is a model interpretation technique introduced by \cite{lundberg2017unified} and inspired by Shapley values, a concept from cooperative game theory to calculate feature importance  $\phi_i(f,x)$. It quantifies feature $i$'s marginal contribution \( \phi_{i} \) to the actual model's prediction for input value $x$. It does so by taking the difference in the model prediction before ($f(S)$) and after ($f(S\cup\{i\})$) feature $i$ is added. Shapley values are weighted and averaged across all possible feature combinations. Formally, the feature importance $\phi_i(f,x)$ is defined as:

\vspace*{-10pt}
\begin{align}
\label{eq:shap}
\phi_i(f,x)=\!\!\!\!\!\sum_{S\subseteq F\setminus\{i\}}\frac{|S|!(F-|S|-1)!}{F!}(f(S\cup\{i\})-f(S))
\end{align}
\vspace*{-10pt}


\noindent where $F$ is the total number of features in the input space and $S$ is the total number of feature coalitions before $i$.  

We calculated SHAP values using Kernel SHAP, which is a computationally efficient, model-agnostic method to approximate SHAP values for any black-box model (including CNNs) using a linear regression model \citep{lundberg2017unified}. We applied Kernel SHAP to segments of the mammographic images obtained via the SLIC (simple linear iterative clustering) algorithm \citep{achanta2012slic} using k-means clustering with 1000 iterations. After evaluating various hyperparameter settings, we selected a configuration of 100 segments with a compactness value of 1, which optimally balances the trade-off between spatial and color proximity. This setup was particularly effective in capturing the edges and shapes of abnormalities in the mammographic images.

\subsubsection{Gradient-weighted Class Activation Mapping (Grad-CAM)}

Grad-CAM is a gradient-based technique to provide visual explanations for a  CNN model by highlighting image regions that are important for the model prediction \citep{selvaraju2017grad}. The algorithm first computes the gradients of the output class score $y^c$ with respect to the feature map activations $A_{ij}^k$ of the $k$-th convolutional layer of the CNN. The gradients $\frac{\partial y^c}{\partial A_{ij}^k}$
describe how much each element of the activation map contributes to the prediction of class $c$.  Averaging all gradients over all spatial locations $(i,j)$  of the feature map, with a normalization factor $Z$ for the total number of pixels in the activation map, yields the neuron importance weights $\alpha_k^c$ assigned to the $k$-th activation map for class $c$. To obtain the Grad-CAM localization map, the weights $\alpha_k^c$ are then used to weight the feature activation maps $A^k$, producing a weighted sum that is passed through a ReLU activation function:

\vspace*{-10pt}
\begin{align}
\text{Grad-CAM}(y^c) &= \text{ReLU}\left(\sum_k\alpha_k^cA^k\right) 
\label{formula:gradcam}
\end{align} 
\vspace*{-10pt}

\noindent where $\alpha_k^c = \frac{1}{Z}\sum_i\sum_j\frac{\partial y^c}{\partial A_{ij}^k}$

As recommended by \cite{selvaraju2017grad}, we used the last convolutional layer ($k = 3$) since it captures high-level features, represented in deeper convolutional layers, while preserving spatial information that is lost in the subsequent fully connected layer. 

\subsubsection{Local Interpretable Model-Agnostic Explanations (LIME)}

LIME suggests local interpretable models that can approximate any complex 'black-box' classifier for a specific data point \citep{ribeiro2016should}. The method produces perturbated samples close to the original data point which are then converted to interpretable representations. 

An explanation for the original data point $x$ is sought given a model $f$ with the predicted class $f(x)$. A local interpretable linear model $g$ is used to approximate $f$, with the proximity measure $\pi_{x_0}$ indicating the similarity between $x$ and the perturbated instances $z$ used to train $g$. LIME aims to minimize the loss $L(f,g,\pi_{x_0})$, which describes how well $g$ approximates $f$ in the neighborhood of $x$ while restricting the model complexity $\Omega(g)$. The regularization term $\Omega(g)$ penalizes complex models. For linear models, for example, complexity is defined as the number of non-zero coefficients. The explanation produced by LIME at a local point $x$ is formally defined as:

\vspace*{-10pt}
\begin{align}
\xi(x) = \arg\min_{g\in G} L(f,g,\pi_{x_0}) + \Omega(g)
\label{formula:lime}
\end{align}
\vspace*{-10pt}

The conceptual intuition between LIME is visualized in \figureref{fig:lime}. The bold star in \figureref{fig:lime} represents the data point $x$ for which an explanation is sought. Red stars are generated samples in the neighborhood of $x$ belonging to the pink class, whereas blue circles describe perturbated samples that belong to the blue class. The size of the perturbated samples represents the proximity to $x$. The complex decision boundary of the model $f$ is depicted as the border of the pink and blue background, whereas the black dashed line is the explainable model $g$ that approximates $f$.


\begin{figure}[htbp]
\floatconts
{fig:lime}
{\captionsetup{aboveskip=-2pt}
\caption{The intuition behind LIME. Adapted from \citet{ribeiro2016should}.}}
{\fbox{\includegraphics[width=.3\textwidth]{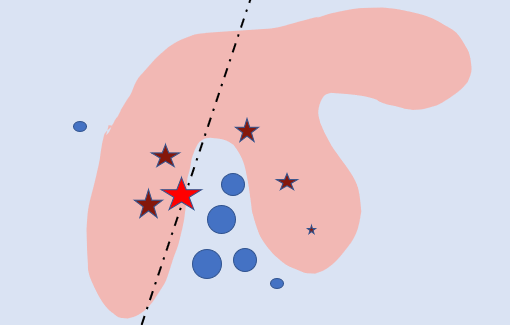}}}
\vspace{-1em} 
\end{figure}

\subsection{Evaluation of Interpretability Algorithms}
We evaluate the ability of LIME, Kernel SHAP, and Grad-CAM to explain the predictions of the CNN model through several criteria. These criteria are 1) computational efficiency, estimated on the basis of running time to determine time complexity under practical time constraints; 2) robustness, assessed by the consistency of explanations for a single input across multiple runs; 3) quality of visual explanations (masks and heatmaps) generated by each algorithm, established through the alignment of explanations with lesioned areas in mammograms.


\section{Results}

\begin{table*}[b!]
\floatconts
  {tab:avg_metrics}
  {\captionsetup{aboveskip=-7pt}
  \caption{Average overall evaluation metrics}}
  {%
    \centering
    \resizebox{\textwidth}{!}{%
    \begin{tabular}{@{}lrrrrr@{}}
    \toprule
    {} & \bfseries Macro Precision &  \bfseries Macro Recall &  \bfseries Macro F1-score & \bfseries  Overall Accuracy &  \bfseries Balanced Accuracy \\
    \midrule
    Majority baseline       &       0.00 & 0.33 & 0.00 & 0.65 & 0.33 \\
    CNN model      &       0.64 & 0.66 & 0.65 & 0.77 & 0.66 \\
    \bottomrule
    \end{tabular}
    }
  }
\end{table*}

Below, we describe the results of the analysis. The result section is divided into two parts. First, we present the performance of the CNN. Second, we evaluate the three interpretability algorithms: Kernel SHAP, LIME, and Grad-CAM.

\subsection{CNN model performance}

As noted above, the CNN model was evaluated on 48 mammographic test images and for a more robust evaluation, the averaged results of multiple independent CNN runs are reported. The overall evaluation metrics for the performance of the CNN model are presented in Table \ref{tab:avg_metrics}. The CNN model outperforms a majority-predicting baseline model achieving an overall accuracy of 0.77. The negligible difference between macro recall (0.66) and macro precision (0.64) suggests a good balance between capturing true positive cases (recall) and correctly classifying positive predictions (precision).

Class-specific performance is presented in Table \ref{tab:class_metrics}. The model achieved the best results for normal cases (Class 0; F1 = 0.93, AUC: 0.98), while the model's performance was least good for malignant cases (Class 2; F1 = 0.44, AUC: 0.83). The benign class was predicted moderately well (Class 1; F1 = 0.58, AUC: 0.90). The class-specific AUC curves are visualized in \figureref{fig:history} in Appendix~\ref{app:ROC}. All class-specific evaluation metrics exceeded baseline performance, indicating that the model learned discriminative patterns across the different classes to at least some extent. An overview of true classes versus predicted classes is presented in the confusion matrix in \figureref{fig:avg_cm}. Overall, the average number of correct predictions was 37.2, whereas the average number of incorrect predictions was 10.8.

\begin{table}[h!]
\floatconts
  {tab:class_metrics}
  {\captionsetup{aboveskip=-7pt}
  \caption{Averaged per-class evaluation metrics}}
  {\centering
    \small
   \begin{tabular}{lrrr}
   \toprule
   {} & \bfseries Class 0 & \bfseries Class 1 & \bfseries Class 2 \\
   \midrule
   Precision &     0.96 &     0.56 &     0.41 \\
   Recall    &     0.90 &     0.60 &     0.48 \\
   F1-score  &     0.93 &     0.58 &     0.44 \\
   Baseline F1-score & 0.79 & 0.00 & 0.00 \\
   \bottomrule
   \end{tabular}
  }
\end{table}

\begin{figure}[htbp]
\floatconts
{fig:avg_cm}
{\captionsetup{aboveskip=-7pt}
 \caption{Averaged confusion matrix.}}
{\includegraphics[width=1\linewidth]{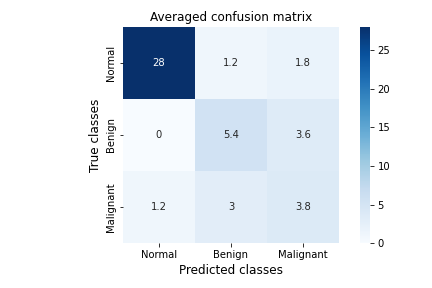}}
\end{figure}

\subsection{Performance of evaluation algorithms}

Below, we present the evaluation of the performance of the interpretability algorithms Kernel SHAP, LIME, and Grad-CAM in terms of computational efficiency, robustness, and quality of visual explanations. 

\subsubsection{Computational efficiency}

The efficiency of each interpretability algorithm was evaluated based on the average running time per image, as presented in Table \ref{tab:time_complexity}. Running times between algorithms differed substantially. Grad-CAM was the fastest algorithm, with an average running time of 0.29 seconds per image. By contrast, LIME required an average of 50.5 seconds per image to compute an explanation.

\begin{table}[htbp]
  \floatconts
    {tab:time_complexity} 
    {\captionsetup{aboveskip=-7pt}
    \caption{Average running time (s) per image for LIME, Grad-CAM, and Kernel SHAP}} 
    {\centering 
    \small 
    \begin{tabular}{lrrr} 
    \toprule
    {} & \bfseries LIME &  \bfseries Grad-CAM &  \bfseries Kernel SHAP \\
    \midrule
    {} &  50.5 &      0.29 &         4.48 \\
    \bottomrule
    \end{tabular}}
\end{table}

\subsubsection{Robustness}

\begin{figure*}[t!]
\floatconts
  {fig:interpretability_stability}
  {\captionsetup{aboveskip=-11pt}
  \caption{LIME (a) and Kernel SHAP (b) both produce different results over multiple runs.}}
  {%
    \subfigure[LIME]
      {\label{fig:lime_stability}
        \includegraphics[width=0.45\linewidth]{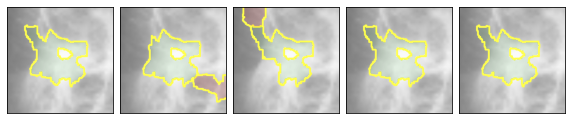}}
      \vspace{1em}
    \subfigure[Kernel SHAP]{
      \label{fig:shap_stability}
        \includegraphics[width=0.45\linewidth]{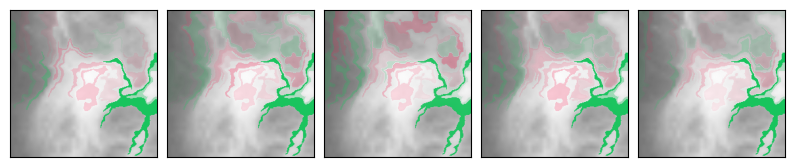}
    }
  }
\end{figure*}

Whereas Grad-CAM is a deterministic algorithm, LIME and SHAP are sampling-based techniques. As a consequence, results for an image vary between different runs of both of these algorithms. The robustness analysis revealed that differences between runs are substantial for LIME. An example of this is presented in 
\figureref{fig:lime_stability}, which visualizes the explanations computed by LIME for five runs on the same image. As can be seen in \figureref{fig:shap_stability}, the differences across runs were more subtle for SHAP.

\begin{figure*}[htbp]
\centering

\begin{minipage}[t]{0.47\textwidth}
    \centering
    \subfigure[LIME (benign)]{
        \includegraphics[width=\linewidth,height=2cm,keepaspectratio]{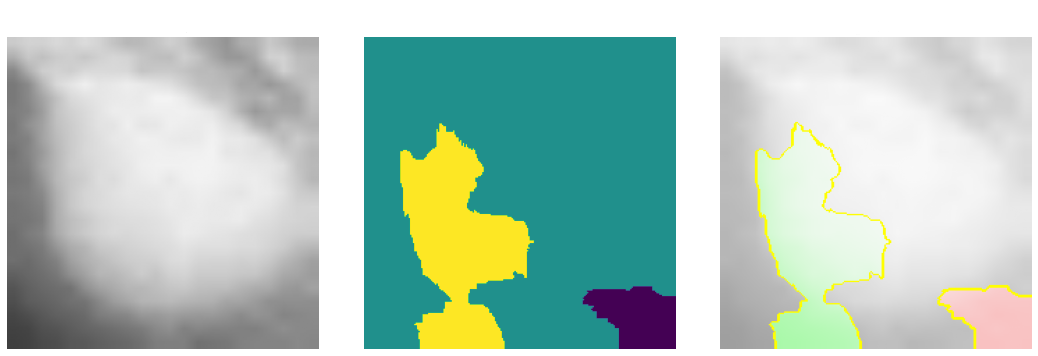}
        \label{fig:true_benign_interpret_lime}
    }

    \subfigure[Kernel SHAP (benign)]{
        \includegraphics[width=\linewidth,height=2cm,keepaspectratio]{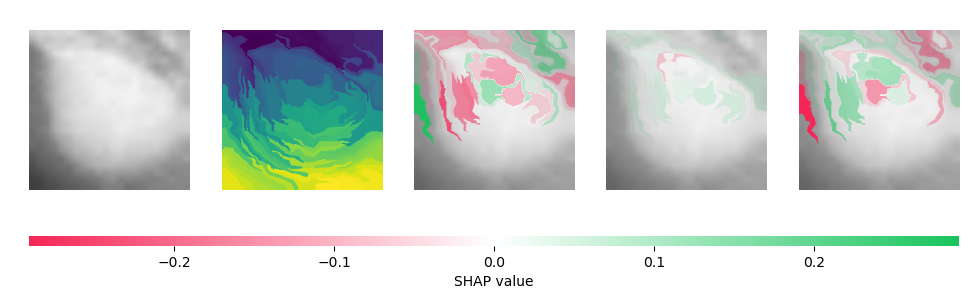}
        \label{fig:true_benign_interpret_shap}
    }

    \subfigure[Grad-CAM (benign)]{
        \includegraphics[width=\linewidth,height=2cm,keepaspectratio]{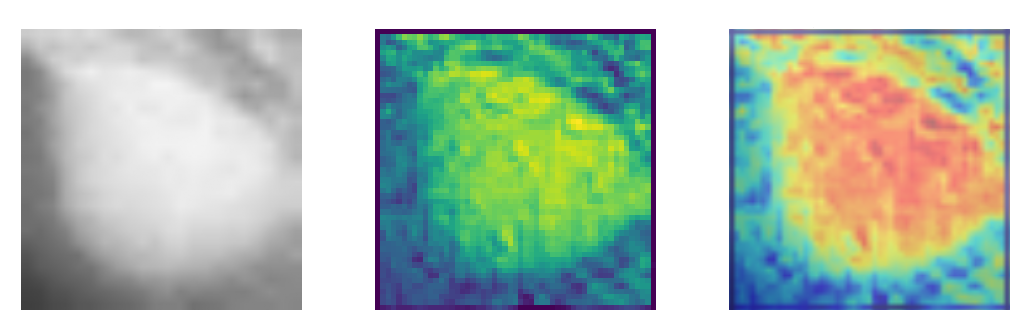}
        \label{fig:true_benign_interpret_gradcam}
    }
\end{minipage}\hfill
\begin{minipage}[t]{0.47\textwidth}
    \centering
    \subfigure[LIME (malignant)]{
        \includegraphics[width=\linewidth,height=2cm,keepaspectratio]{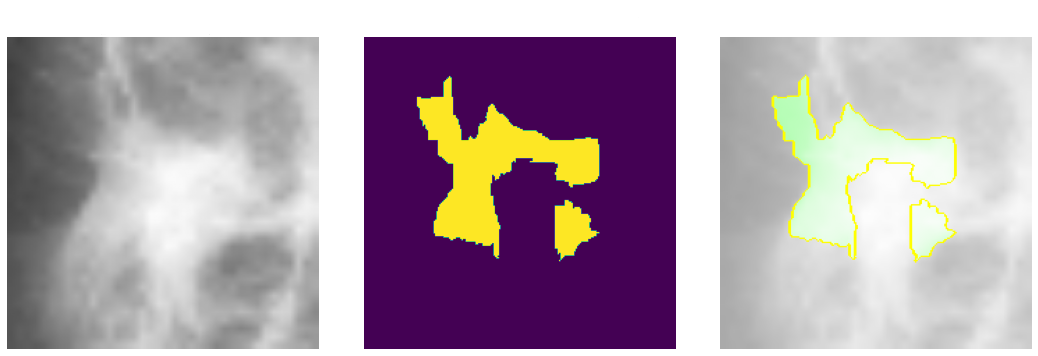}
        \label{fig:true_pred_malignant_interpret_lime}
    }

    \subfigure[Kernel SHAP (malignant)]{
        \includegraphics[width=\linewidth,height=2cm,keepaspectratio]{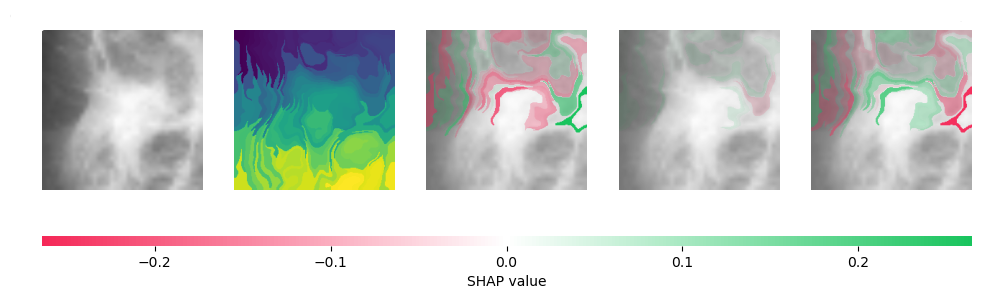}
        \label{fig:true_pred_malignant_interpret_shap}
    }

    \subfigure[Grad-CAM (malignant)]{ \includegraphics[width=\linewidth,height=2cm,keepaspectratio]{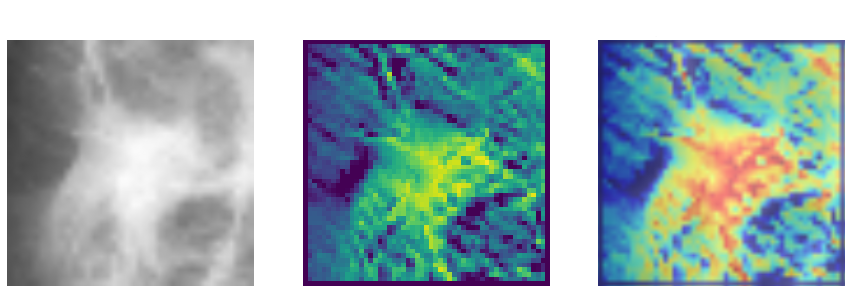}
        \label{fig:true_pred_malignant_interpret_gradcam}
    }
\end{minipage}
\caption{Original image (left of panel), heatmap (middle of panel) and overlaid heatmap (right of panel) for the explanations of the LIME (a, d), 
Kernel SHAP (b, e), and Grad-CAM (c, f) algorithms for a benign (left) and malignant (right) mammogram.} 
\label{fig:interpretation_plots}
\end{figure*}

\subsubsection{Comparison of visual explanations}

The effectiveness of LIME, Kernel SHAP, and Grad-CAM was examined for a correctly classified benign mammogram, with typical oval shape and smooth borders, and for a malignant abnormality, with characteristic irregular, tentacle-like patterns. The corresponding explanations are 
visualized in \figureref{fig:interpretation_plots}.

LIME, highlighting positively and negatively contributing areas to the prediction in green and red, revealed partial coverage with the lesions only. Kernel SHAP offers separate explanations for benign (third image in panel), malignant (fourth image in panel), and normal (fifth image in panel) predictions - using green and red to signify superpixels that increase or decrease class probabilities. As can be seen in \figureref{fig:interpretation_plots}, 
it succeeded in identifying key edges in both benign (Figure \ref{fig:true_benign_interpret_shap}) and malignant abnormalities (Figure \ref{fig:true_pred_malignant_interpret_shap}) that overlap with the lesion borders. Grad-CAM, employing a red-to-blue heatmap, highlights pixels that were the most and least influential in the classification process. The explanations of Grad-CAM revealed significant overlap with the actual 
shapes of both benign (\figureref{fig:true_benign_interpret_gradcam}) and malignant abnormalities (Figure \ref{fig:true_pred_malignant_interpret_gradcam}) and were closely aligned with the characteristics of the lesion. Appendix \ref{fig:more_interpret_visualizations_incorrect} presents the explanations for LIME, Kernel SHAP, and Grad-CAM for an incorrect prediction (i.e., a malignant image classified as benign).  

\begin{figure*}[htbp]
\floatconts
  {fig:overall_gradcam_predictions} 
  { \captionsetup{aboveskip=-3pt}
  \caption{Grad-CAM predictions. Red boxes indicate incorrect predictions.}} 
  {%
    \subfigure[Normal predictions]{%
      \label{fig:overall_normal_gradcam_predictions} 
      \includegraphics[width=0.9\linewidth]{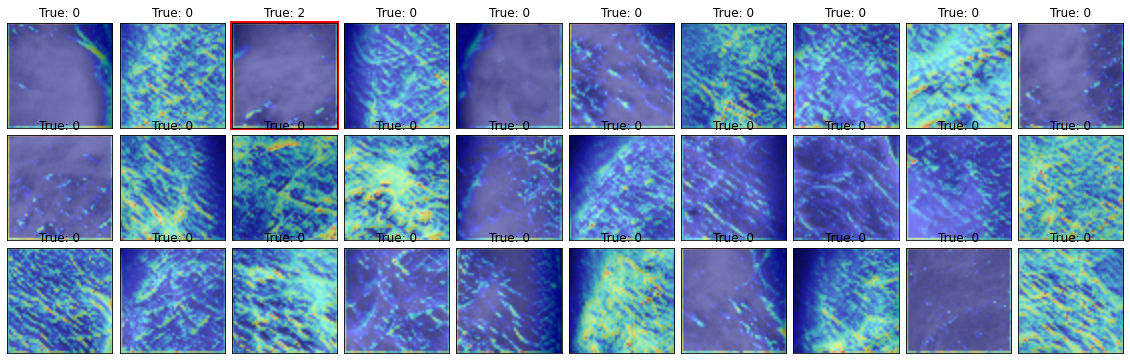}
    }\vspace{0.1cm} 
    \subfigure[Benign predictions]{%
      \label{fig:overall_benign_gradcam_predictions} 
      \includegraphics[width=0.9\linewidth]{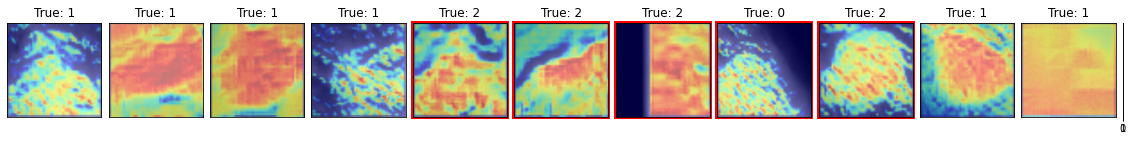}
    }\vspace{0.1cm} 
    \subfigure[Malignant predictions]{%
      \label{fig:overall_malignant_gradcam_predictions} 
      \includegraphics[width=0.9\linewidth]{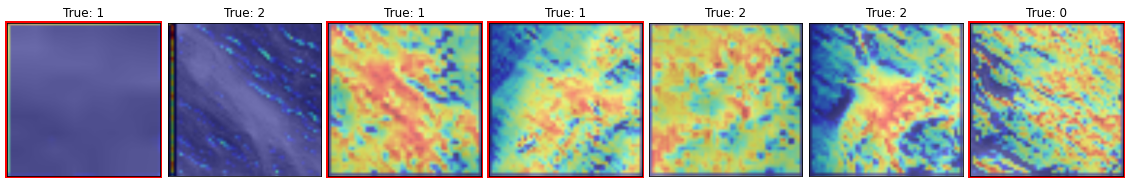}
    }
  }
\end{figure*}

Given its excellent performance, we present the explanations offered by the Grad-CAM algorithm in more detail in \figureref{fig:overall_gradcam_predictions}, which shows Grad-CAM explanations for all mammographic images in the dataset, categorized by predicted class. Images for which the predictions of the CNN were incorrect are marked by a red border. Analogous figures for LIME (\figureref{fig:LIME_predictions}) and Kernel SHAP (\figureref{fig:KernelSHAP_predictions}) are presented in Appendix \ref{fig:more_interpret_visualizations_lime_kernel_shap}. Explanations for normal predictions are characterized by a lack of distinct patterns (\figureref{fig:overall_normal_gradcam_predictions}), explanations for true benign cases by round or oval shapes (\figureref{fig:overall_benign_gradcam_predictions}), and malignant cases - with two notable exceptions - by more irregular, undefined shapes (\figureref{fig:overall_malignant_gradcam_predictions}). As such, Grad-CAM is able to differentiate between the three classes of images effectively.

\section{Discussion} 

We presented a study of the model interpretation algorithms LIME, Grad-CAM, and Kernel SHAP in the context of a CNN fit to the MIAS dataset for mammographic image classification. Below, we discuss the findings of this study in more detail. The discussion is divided into four sections: the evaluation of the CNN, the evaluation of the explanations offered by the three interpretation algorithms that help gain insight into the CNN's 'black-box' model predictions, practical implications for clinical relevance and directions for future research.

\subsection{Evaluation of the CNN}

We applied a variety of preprocessing techniques to enhance the MIAS mammogram dataset's usability for CNN classification, including noise removal, image contrast enhancement, and ROI extraction, supplemented by data augmentation addressing the original dataset size limitations. To reduce model bias towards the majority class, the classes were balanced. This preprocessed dataset is now publicly available, promoting research transparency and alleviating time-intensive preprocessing requirements. However, the dataset's inherent limitations, including low image quality and the absence of a second breast view potentially negatively affected the performance of the CNN. The reliance on manual ROI annotations and classification labels, while necessary, also introduces potential biases, as mammogram interpretation is subject to human error, especially for rare cancer types that have been shown to remain undetected by radiologists \citep{evans2013if}. Future research could explore the fine-tuning of the trained ROI model on whole mammogram images as suggested by \cite{shen2019deep}, to enable the classification of complete mammogram images in the absence of ROI annotations. Given sufficient computational resources, mammograms that offer higher resolution and two breast views such as the INBreast \citep{moreira2012inbreast} or DDSM dataset, \citep{heath1998current}, could therefore provide finer-grained information to improve the classification performance \citep{petrini2022breast}.

The CNN effectively distinguished between normal, benign, and malignant cases with few misclassified normal cases (recall = 0.90). However, it showed a higher prevalence of misclassified malignant lumps (recall = 0.48), leaving more than half of true cancer cases undetected. This highlights the need for improved malignancy detection - as false negatives can have severe negative implications such as treatment delays or a false sense of security as the women are unaware of their disease \citep{nci_facts}. As the majority of incorrect classifications involved the distinction between benign and malignant cases, future studies should consider a more fine-grained classification between these two classes, such as a five-class classification, distinguishing calcifications from masses \citep{shen2019deep}. Additionally, deploying deeper CNNs with smaller kernel sizes \citep{li2019benign} and exploring more advanced architectures such as Vision Transformers \citep{dosovitskiy2020image} could further reduce false positives. 

Our use of an averaged confusion matrix mitigated the influence of variations across different training runs, thus enhancing metric reliability by capturing a more general trend. Yet, training and evaluating the model on a single data split may have biased the results towards that specific split. While our findings suggest that the CNN identifies characteristic tumor shapes, we acknowledge our dataset's limited diversity and stress the importance of further evaluations on broader datasets. Such assessments are crucial to verify the CNN's capability to accurately distinguish between various breast cancer types, including rare and complex cases, across different patient populations and screening conditions. 
Furthermore, the approach as suggested by \cite{agarwal2019automatic} could be adapted to not only test but fine-tune the model by training on multiple mammographic datasets, potentially improving its performance and making the model more robust to nuances present in different datasets.

\subsection{Evaluation of explanations}

The main objective of the current study was to provide better insight into the reasons why the CNN classified a mammogram as normal, benign, or malignant. Specifically, we compared the algorithms Grad-CAM, SHAP, and LIME with respect to their computational efficiency, robustness, and quality of explanations. When adding an interpretation algorithm to a breast cancer detection procedure it is of vital importance that the algorithm offers explanations that are computationally efficient as well as consistent. Time-efficient explanations are crucial because radiologists face time constraints during clinical decision-making. Additionally, trust into an AI-powered technology is an important factor for clinicians to adopt it in practice \citep{tucci2022factors}. Consistency of an interpretation algorithm, therefore, is essential \citep{yu2013stability,lee2019developing}.

In light of these considerations, LIME - which takes an average of 50.5 seconds to compute an explanation and generates highly unstable masks - may be less suitable in clinical practice. SHAP, similar to LIME, is a perturbation-based approach that involves the generation of new data points as part of the explanation process and thus renders explanations in a non-deterministic manner. Nonetheless, SHAP explanations revealed less variability as compared to LIME explanations. Furthermore, SHAP was more computationally efficient, as indicated by reduced running times. Optimal performance in terms of computational efficiency and robustness, however, was observed for Grad-CAM, which takes less than one second (0.29 seconds) per image to generate explanations and provides deterministic explanations by directly utilizing the model gradients. Grad-CAM thus appears to be highly suitable for practical implementation.

Benign and malignant cancer appears as white or light-grey matter on mammograms and thus sets itself apart from the grey breast tissue. Based on this understanding, the assumption was made that benign and malignant abnormalities are characterized by white lumps on mammograms, disregarding the background. LIME and Kernel SHAP were unable to accurately identify the entire breast lesion. By contrast, Grad-CAM identified lesions (most)  accurately and generated (the most) plausible explanations by effectively identifying distinctive features corresponding to normal, benign, and malignant breast tissue characteristics. The Grad-CAM heatmaps for benign predictions showed clearly defined round/oval shapes, indicating the presence of non-spreading, non-cancerous abnormal tissue. The heatmaps for malignant predictions displayed less-defined, scattered areas spreading into different directions, aligning with the invasive characteristic of cancerous malignant cells into the surrounding areas. Normal predictions did not exhibit distinctive patterns, consistent with the expectation that healthy breast tissue does not exhibit suspicious abnormalities. As such, the explanations offered by Grad-CAM fit well with 
the patterns observed in the diagnosis of breast cancer abnormalities in 
clinical practice \citep{gco}. 

\subsection{Practical Implications for Clinical Relevance}
The suitability of Grad-CAM in providing explanations aligning with human intuitions have practical implications for the clinical practice. We propose a system augmenting a CNN-based classifier with visual explanations in the form of heatmaps, serving as a foundational tool for future research by incorporating interpretability within automated breast cancer classification. This does not only enhance transparency of DL-based automated mammography diagnosis, but can furthermore serve as an educational tool for medical trainees by providing large number of examples of characteristic heatmaps of breast cancer types. Additionally, our framework could be extended to not only identify well-known features of known breast abnormalities but also uncover novel diagnostic markers of rare cancer types.

\subsection{Future Directions}

We presented an initial exploration of interpretation algorithms in the context of mammography image classification. In this initial exploration,
we focused on the qualitative evaluation of different interpretability algorithms, acknowledging the possibility of observer bias. Future research, inspired by previous experiments conducted by \cite{ribeiro2016should} and \cite{selvaraju2017grad} could and should involve human-subject experiments to more reliably assess the extent to which Grad-CAM enhances trust of radiologists in deep learning-based technologies. Additionally, further investigation is in order to evaluate to what extent Grad-CAM is effective in detecting human diagnostic errors, particularly in identifying misclassified benign or malignant cases and in the identification of rare cancer types that may be missed by radiologists \citep{evans2013if}. Furthermore, it is important to note that we only compared post-hoc methods in the current study. Post-hoc methods are approximations of the original model and don't modify the 'black-box' architecture of the CNN model itself. The use of 'white-box' models, intrinsically interpretable CNNs \cite[see, e.g.][]{zhang2018interpretable}, is another avenue to explore in future research.

Finally, evaluating interpretability techniques still relies on the expertise and judgment of the observer, as there are no standardized qualitative or quantitative measures to assess the plausibility of provided explanations. For future research, we therefore recommend the involvement of domain experts, such as professional radiologists, to establish a ground truth. More specifically, since the ROIs in the MIAS dataset lack exact lesion masks, professional radiologists could mark the precise breast abnormalities present in the ROI for a quantitative analysis of the predictions. This would enable the calculation of overlap between the interpretation mask and annotations provided by expert radiologists, using metrics like intersection over union (IoU) to measure the agreement between both.

\section{Conclusions}

We presented the application of three interpretation techniques - LIME, Kernel SHAPE, and Grad-CAM - to the results of a CNN trained on mammographic image data. Grad-CAM emerged as the preferred interpretation technique for the current data set and machine learning model, providing insightful explanations of the model's predictions in a time-efficient and stable way. The current findings provide valuable insights into the interpretability of deep learning models in the context of breast cancer classification and demonstrate the potential of explainable artificial intelligence (XAI) as a supplementary tool for radiologists in the diagnosis of breast cancer.


\nolinenumbers

\bibliography{references}
\clearpage

\onecolumn
\appendix

\setcounter{table}{0}
\renewcommand{\thetable}{A\arabic{table}}
\setcounter{figure}{0}

\renewcommand{\thefigure}{A\arabic{figure}}


\section{CNN architecture}\label{app:CNN architecture}
\begin{table*}[h!]
\floatconts
  {tab:cnn_architecture}
  {\caption{CNN Architecture adapted from \cite{el2021malignant}. A dropout layer was added after the third fully-connected layer for model regularization purposes and the originally binary problem was transformed into a 3-class classification problem.}}
  {\resizebox{\textwidth}{!}{%
    \begin{tabular}{@{}ccccccccc@{}}
    \toprule
  \bfseries  Layer \# &  \bfseries Kernels &  \bfseries Kernel size &  \bfseries Stride &  \bfseries Padding &  \bfseries Output Shape &  \bfseries Output Size & \#  \bfseries Parameters &  \bfseries Activation \\
    \midrule
    Input Image & - & - & - & - & 224,224,3 & - & - & - \\
    convo\_1 & 16 & 5 $\times$ 5 & 1 $\times$ 1 & 0 $\times$ 0 & 220,220,16 & 665,856 & 1216 & ReLU \\
    max\_pooling\_1 & - & 2 $\times$ 2 & 2 $\times$ 2 & - & 110,110,16 & 166,464 & 0 & - \\
    convo\_2 & 16 & 5 $\times$ 5 & 1 $\times$ 1 & 0 $\times$ 0 & 106,106,16 & 153,664 & 6416 & ReLU \\
    max\_pooling\_2 & - & 2 $\times$ 2 & 2 $\times$ 2 & - & 53,53,16 & 38,416 & 0 & - \\
    convo\_3 & 14 & 3 $\times$ 3 & 1 $\times$ 1 & 1 $\times$ 1 & 53,53,14 & 33,614 & 2030 & ReLU \\
    max\_pooling\_3 & - & 2 $\times$ 2 & 2 $\times$ 2 & - & 26,26,14 & 8064 & 0 & - \\
    max\_pooling\_4 & - & 2 $\times$ 2 & 2 $\times$ 2 & - & 13,13,14 & 2016 & 0 & - \\
    flatten\_1 & - & - & - & - & - & 2016 & 0 & - \\
    dense\_1 & - & - & - & - & 512,1 & 512 & 1,211,904 & - \\
    dense\_2 & - & - & - & - & 256,1 & 256 & 131,328 & - \\
    dense\_3 & - & - & - & - & 128,1 & 128 & 32,896 & - \\
    dropout\_1 & - & - & - & - & 512,1 & - & - & - \\
    dense\_4 & - & - & - & - & 3,1 & 3 & 387 & - \\
    \midrule
    Total params: & - & - & - & - & - & - & 1,386,177 & - \\
    \bottomrule
    \end{tabular}%
  }}
\end{table*}

The model architecture consists of three convolutional layers, each followed by max-pooling layers (two max-pooling layers after the third convolutional layer), with strides of 2 for pooling and 1 for convolution. Kernel sizes are 5x5 for the first two layers and 3x3 for the last, with 16, 16, and 14 kernels, respectively. The network incorporates ReLU activation functions, He weight initialization \citep{he2015delving}, three fully connected layers with 512, 256, and 128 units after the flattening layer, followed by a final softmax function for three-class (normal, benign, malignant) classification.

\newpage

\section{ROC curve} \label{app:ROC}

\begin{figure}[htbp]
\floatconts
{fig:history}
{\captionsetup{aboveskip=-2pt}
\caption{ROC curve for normal (blue), benign (orange), and malignant (green) predictions.}}
{\includegraphics[width=.5\textwidth]{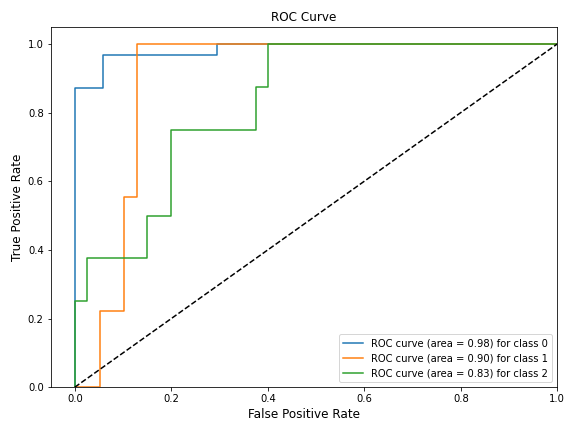}}
\vspace{-1em} 
\end{figure}

 \section{Explanations for an incorrect prediction} \label{fig:more_interpret_visualizations_incorrect}

\begin{figure}[htbp]
\floatconts
  {fig:false_malignant_interpret}
  {\caption{Original image (left of panel), heatmap (middle of panel) and overlaid heatmap (right of panel) for the explanations of the LIME (a), 
Kernel SHAP (b), and Grad-CAM (c) algorithms for a malignant mammograms that was incorrectly predicted to be benign.}}
  {%
    \begin{minipage}[t]{0.45\textwidth}
        \centering
        \subfigure[LIME]{
            \includegraphics[width=\linewidth]{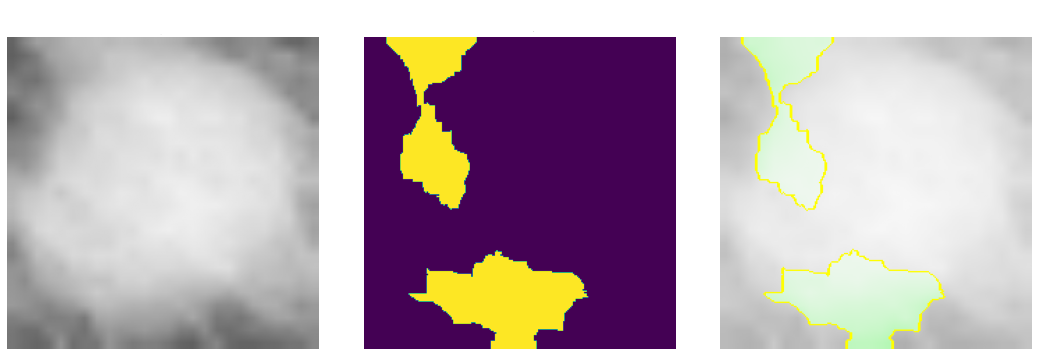}
            \label{fig:false_pred_interpret_lime}
        }
        \subfigure[Kernel SHAP]{
            \includegraphics[width=\linewidth]{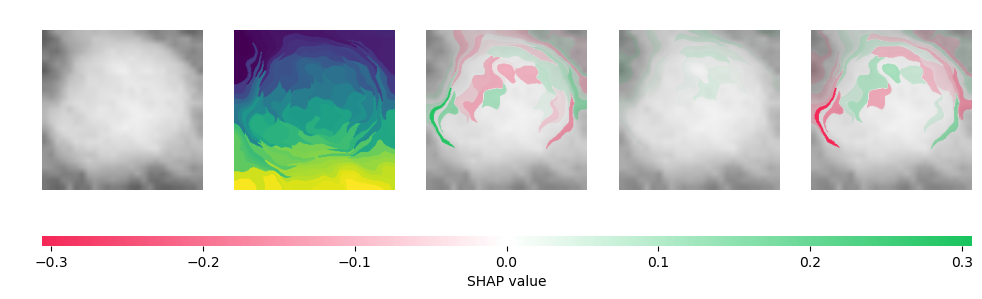}
            \label{fig:false_pred_interpret_shap}
        }
        \subfigure[Grad-CAM]{
            \includegraphics[width=\linewidth]{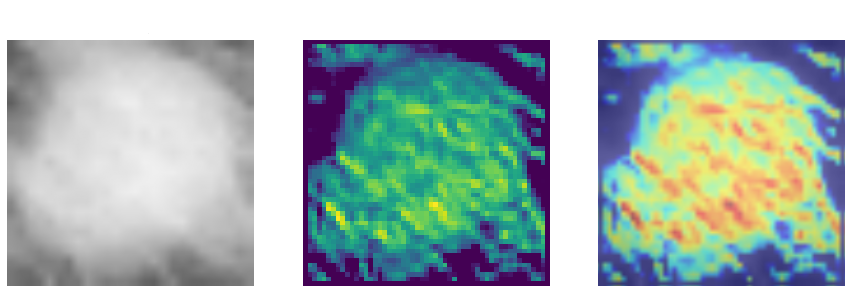}
            \label{fig:false_pred_interpret_gradcam}
        }
    \end{minipage}
  }
\end{figure}

\newpage

 \section{Predictions LIME and Kernel SHAP} \label{fig:more_interpret_visualizations_lime_kernel_shap}

\begin{figure}[htbp]
\floatconts
  {fig:LIME_predictions} 
  { \caption{LIME predictions. Red boxes indicate incorrect predictions.}} 
  {%
    \subfigure[Normal predictions]{%
      \label{fig:lime_normal} 
      \includegraphics[width=0.9\linewidth]{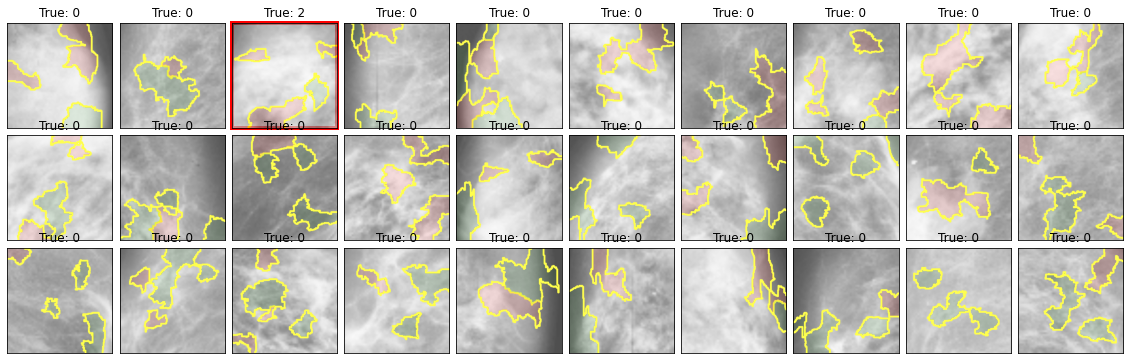}
    }\vspace{0.1cm} 
    \subfigure[Benign predictions]{%
      \label{fig:lime_benign} 
      \includegraphics[width=0.9\linewidth]{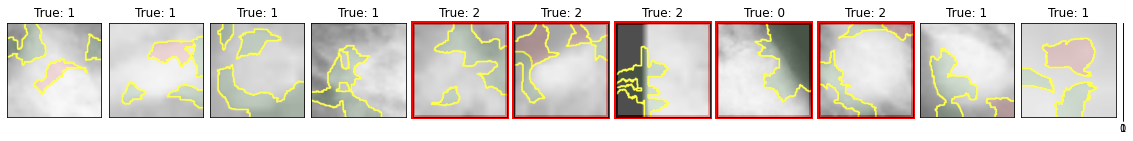}
    }\vspace{0.1cm} 
    \subfigure[Malignant predictions]{%
      \label{fig:lime_malignant} 
      \includegraphics[width=0.9\linewidth]{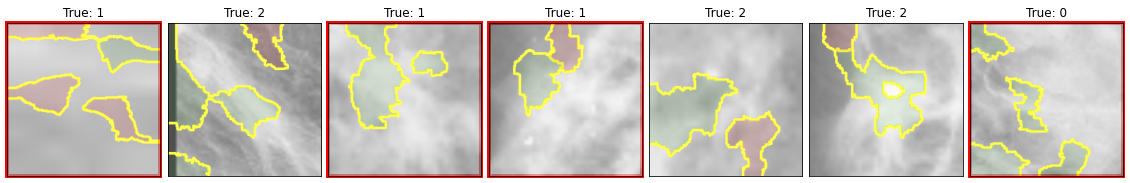}
    }
  }
\end{figure}

\begin{figure}[htbp]
\floatconts
  {fig:KernelSHAP_predictions} 
  { \caption{Kernel SHAP predictions. Red boxes indicate incorrect predictions.}} 
  {%
    \subfigure[Normal predictions]{%
      \label{fig:normal_kernelshap_predictions} 
      \includegraphics[width=0.9\linewidth]{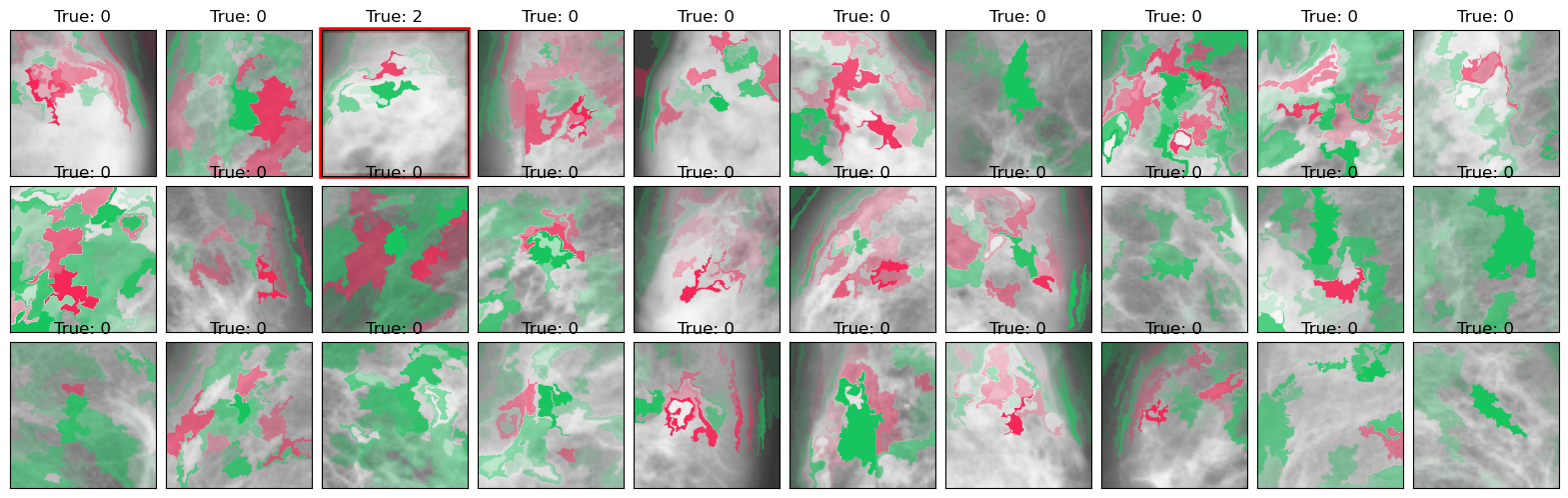}
    }\vspace{0.1cm} 
    \subfigure[Benign predictions]{%
      \label{fig:benign_kernelshap_predictions} 
      \includegraphics[width=0.9\linewidth]{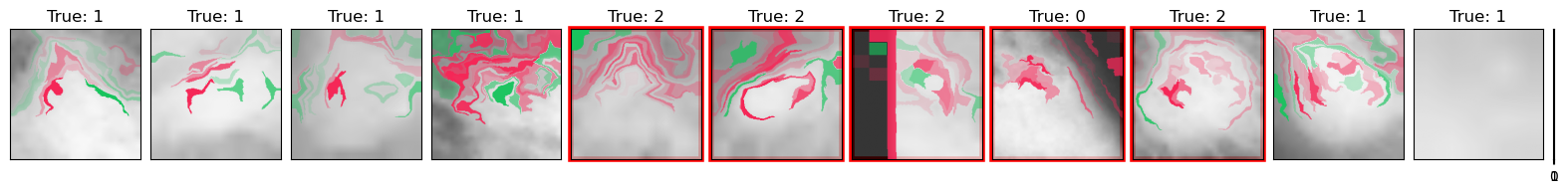}
    }\vspace{0.1cm} 
    \subfigure[Malignant predictions]{%
      \label{fig:malignant_kernelshap_predictions} 
      \includegraphics[width=0.9\linewidth]{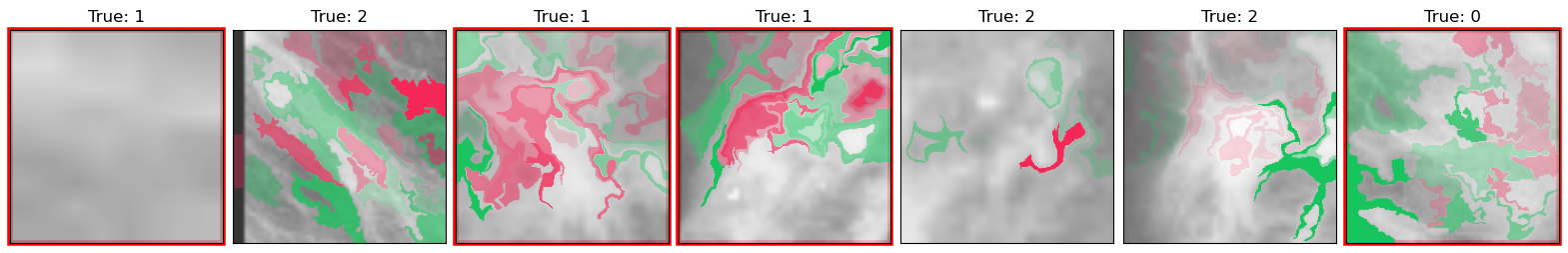}
    }
  }
\end{figure}

\end{document}